\def\year{2020}\relax
\documentclass[letterpaper]{article} 
\usepackage{aaai20}  
\usepackage{times,color}  
\usepackage{helvet} 
\usepackage{courier}  
\usepackage[hyphens]{url}  
\usepackage{graphicx} 
\usepackage{amsmath}
\usepackage{amssymb}
\usepackage{amsopn}
\usepackage{tabularx, booktabs}
\usepackage{mathtools}
\usepackage{algorithmic}
\usepackage[ruled,vlined,linesnumbered,lined,boxed,commentsnumbered]{algorithm2e}
\newtheorem{mydef}{Definition}
\usepackage{graphicx}  
\title{Zero Shot Learning with the Isoperimetric Loss}
\author{Shay Deutsch\textsuperscript{\rm 1}, Andrea Bertozzi\textsuperscript{\rm 1}, and Stefano Soatto\textsuperscript{\rm 2} \\  
\textsuperscript{\rm 1} Department of Mathematics, \textsuperscript{\rm 2} Department of Computer Science \\ 
University of California, Los Angeles\\  
shaydeu@math.ucla.edu\textsuperscript{\rm 1}, bertozzi@math.ucla.edu \textsuperscript{\rm 1} , soatto@cs.ucla.edu  \textsuperscript{\rm 2}
}
\begin{document}

\maketitle

\begin{abstract}
 We introduce the isoperimetric loss as a regularization criterion for learning the map from a visual representation to a semantic embedding, to be used to transfer knowledge to unknown classes in a zero-shot learning setting. We use a pre-trained deep neural network model as a visual representation of image data, a Word2Vec embedding of class labels, and linear maps between the visual and semantic embedding spaces. However, the spaces themselves are not linear, and we postulate the sample embedding to be populated by noisy samples near otherwise smooth manifolds. We exploit the graph structure defined by the sample points to regularize the estimates of the manifolds by inferring the graph connectivity using a generalization of the isoperimetric inequalities from Riemannian geometry to graphs. Surprisingly, this regularization alone, paired with the simplest baseline model, outperforms the state-of-the-art among fully automated methods in zero-shot learning benchmarks such as AwA and CUB. This improvement is achieved solely by learning the structure of the underlying spaces by imposing regularity. 
\end{abstract}
\section{Introduction}

\begin{figure}[htb]
\centering\includegraphics[width=6cm]{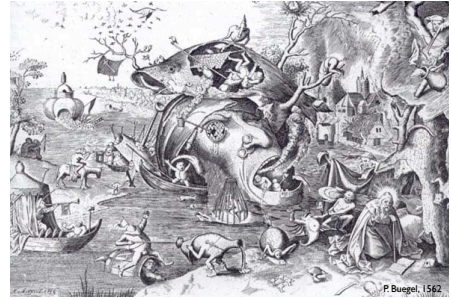}
\caption{Find the Pottopod (Courtesy of P. Perona).}
\label{fig:pottopod}
\end{figure}

\noindent{\textbf{Motivating example}.} A pottopod is a pot with limbs. Not even a single example image of a pottopod is needed to find one in Fig. \ref{fig:pottopod}. However, one has surely seen plenty of examples of animals with limbs, as well as pots. In zero-shot learning (ZSL) one aims to exploit models trained with supervision, together with maps to some kind of attribute or ``semantic'' space, to then recognize objects as belonging to classes for which no previous examples have ever been seen.

\begin{figure}[htb]
  \centering
 \includegraphics[width=5cm]{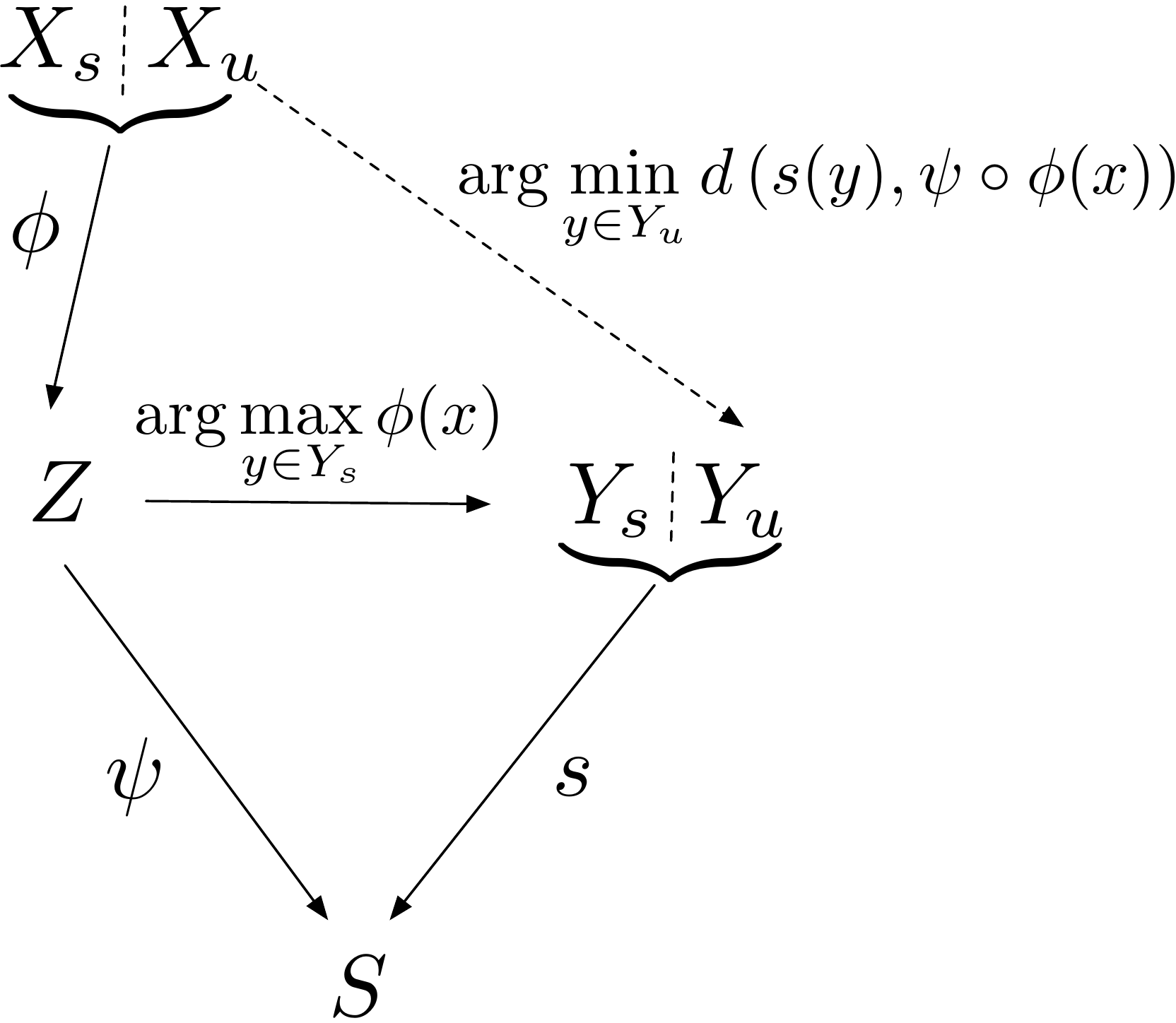}
\caption{Illustration of the components of a ZSL algorithm: For images in the set with seen labels $X_s$, the labels can be estimated by maximum a-posterior over labels in the seen set $Y_s$ on the visual representation $\phi(X_s)$. For the unseen labels, there is no direct connection to the data because they are not seen during training. Inference is then indirect: A visual representation is inferred, and from there a semantic representation, which is compared to the semantic representation of unseen labels, minimizing {\em some} distance in the semantic space, over all possible unseen labels $Y_u$.}
\label{fig:diagram}
\end{figure}

The ingredients of a ZSL method are illustrated in Fig. \ref{fig:diagram}. Seen samples $X_s$ and their corresponding labels $Y_s$ are used to train a model $\phi$, typically a deep neural network (DNN), in a supervised fashion, to yield a vector in a high-dimensional (visual embedding) space $Z$. At the same time, a function $s$ maps semantic attributes such as ``has legs,'' ``is short,'' or simply a word embedding of the ground truth (seen and unseen) labels of interest $Y = \{Y_s, Y_u\}$, to some metric space. The name of the game in ZSL is to learn a map $\psi$, possibly along with other components of the diagram, from the visual embedding space $Z$ to the semantic space $S$, that can serve to transfer knowledge from the seen labels (reflected in $Z$) to the unseen ones.
Alternatively, one can try to cluster samples with unseen labels in the visual embedding space $Z$, and then associate clusters to unseen labels.

\noindent{\bf Focus on regularization.} Transfer of knowledge hinges on some kind of regularity of the maps involved in ZSL. In practice, the visual embedding space $Z$ and the semantic embedding $S$ are only known through discrete samples, and the maps are learned restricted to these finite samples. One crucial theme in ZSL is, interpreting sample embeddings as ``noisy points'' on the otherwise differentiable manifolds $Z$ and $S$, to attempt to regularize the spaces $Z$, $S$, and/or the map between them.

\noindent{\bf Key contribution.} Of all the various components of a ZSL method, we choose the simplest possible \cite{EZSL}, except for the regularization of the semantic map. There, we introduce a sophisticated model, based on an extension of the isoperimetric inequalities from Riemannian geometry to discrete structures (graphs). We treat sample visual embeddings as vertices in a graph, with affinities as edges, and the visual-to-embedding map $\psi$ interpreted as a linear function on the graph. We then introduce the isoperimetric loss (IPL) to enforce regularity on the domain $Z$ based on the flow of the function defined on it through the boundary of sets of a given volume. The resulting regularized graph is informed by both the visual and semantic maps. We use it to perform clustering and map clusters to labels. Therefore, we take a very simple visual-to-semantic embedding function, namely a linear transformation, and indirectly regularize it by regularizing its domain and range spaces. 

We expected our regularization to  improve the baseline \cite{EZSL} on which our ZSL model is based. We did not expect it to surpass the (far more sophisticated) state-of-the-art in the two most common benchmark datasets used in ZSL, namely AwA \cite{Lampert09} and CUB \cite{Cubdataset}. Yet, the experiments indicate so. In some cases, it even outperformed methods that used human annotation for the unseen labels.

At heart, we solve a topology estimation problem. We determine the connectivity between nodes of the visual embedding graph, which defines a topology in that space informed by the semantic representation of seen attributes. Much of the literature in this area focuses on what kind of graph signal (embedding, or descriptor) to attribute to the nodes, whereas the connectivity of the graph is decided a-priori. We focus on the complementary problem, which is to determine the graph connectivity and learn the graph weights. Unlike other approaches, the connectivity in our method is informed both by the value of the  visual descriptors at the vertices, and the values of the semantic descriptors in the range space. Our framework allows us to use automated semantic representation to perform ZSL, resulting in a framework which is entirely free of human annotation.

\section{Preliminaries}
\label{sec:preliminaries}

We first describe a general formalization of ZSL that helps place our contribution in context. Every ZSL includes a supervised component, which results in a visual embedding, a collection of unseen labels or attributes, a map from these attributes to a vector (semantic) space, and a map from visual to semantic spaces. It is important to understand the assumptions underlying the transfer of information from seen to unseen attributes, which translates in regularity assumptions on the visual-to-semantic map.\\
{\bf Supervised component.} In standard supervised classification, a dataset ${\cal D}_s$ is given where both the input data $x$ and the output labels $y_s$ are {\em seen}:
\begin{equation}
    {\cal D}_s = \{x^i, y_s^i\}_{i= 1}^{N}
\end{equation}where the set of seen labels, for instance 1 = ``cat'' and 2 = ``dog,'' is indicated by $Y_s$, with cardinality $|Y_s| = n_s$.  The data belong to $X$, for instance the set of natural images. The goal of supervised learning is to infer the parameters $w$ of a function $\phi_w: X \rightarrow {\mathbb R}_+^{K}$ that approximates the (log)-posterior probability over $Y_s$, 
\begin{equation}
\phi_w(x)_j \simeq \log P(y_s = j | x).
\end{equation}where the subscript $j$ denotes the $j$-th component of the vector $\phi_w(x)$. At test time, given an unseen image $x$, one can infer the unknown label $\hat y$ associated with it as the maximum a-posteriori estimate
\begin{equation}
\hat y(x) = \arg\max_{y \in Y_s} \phi_w(x)_y.
\end{equation}
We indicate with $Z$ the (latent, or representation) space where the data $X$ are mapped, 
\begin{equation}
     z^i = \phi_w(x^i)  \in Z.
\end{equation}{\bf Visual embedding.} Although $z^i$ can be interpreted as log-probabilities, one can simply consider them as an element of a vector space of dimension at least $K$, $Z \subset {\mathbb R}^K$, called ``visual embedding.'' It is also customary to use intermediate layers of a deep network, rather than the last one that is used for classification, as a visual embedding, so in general $K \neq n_s$. A general classifier, rather than a linear one, can then be used to determine the class, based on the latent representation $z$. We want the formalism to be flexible, so we do not constrain the dimension of the embedding to be the same of the dimension of the seen classes, and we consider $z = \phi_w(x)$ to be any (pre-trained) visual feature.

\noindent {\bf Unseen labels.} In ZSL there is a second set of ``unseen'' labels\footnote{`A misnomer, since one knows ahead of time what these labels are, for instance 3 = ``sailboat'', 4 = ``car.'' However, one is not given images with those labels during training. So, while the labels are seen, image samples with those labels are not seen during training. } $Y_u$, disjoint from the first $Y_u \cap Y_s = \emptyset$. We call $Y = Y_u \cup Y_s$. At training time we do not have any sample images with labels in $Y_u$. However, we do at test time. 

\noindent{\bf Zero-shot learning.} The goal of ZSL is to classify test images as belonging to the unseen classes. That is, to learn a map from $X$ to $Y_u$. Absent any assumption on how the unseen labels are related to the seen labels, ZSL makes little sense.

\noindent{\bf Assumptions.} In ZSL one assumes there is a ``semantic'' metric vector space $S \subset {\mathbb R}^M$, to which all labels -- seen and unseen -- can be mapped via a function $s: Y \rightarrow S$. If the metric is Euclidean, a distance between two labels, $y^i, y^j$, can be induced via $d_s(y^i, y^j) = \| s(y^i) - s(y^j) \|.$
Otherwise, any other distance $d(s(y^i), s(y^j))$ on $S$ can be used to find the label associated to an element $\sigma \in S$ of the semantic space (embedding), for instance using a nearest-neighbor rule
\begin{equation}
    \hat y (\sigma) = \arg\min_{y \in Y} d(s(y) - \sigma).
\end{equation}Note that the minimum could be any label, seen or unseen. This is just a metric representation of the set of labels, independent of the ZSL problem.

The second assumption is that there exists a map $\psi: Z \rightarrow S$ from the visual to the semantic embedding, which can be learned to map embeddings of seen classes $z$  to semantic vectors $\sigma = \psi(z)$ in such a way that they land close to the semantic embedding $s(y_s)$ of the seen labels:
\begin{equation}
     \arg\min_\psi \sum_{i=1}^N d(s(y^i_s) - \psi \circ \phi_w (x^i_s)).
    \label{eq-ZSL-loss}
\end{equation}One could learn both the visual embedding $\phi_w$ and the visual-to-semantic map $\psi$ simultaneously, or fix the former and just learn the latter. In some cases, even the latter is fixed. 

\noindent{\bf Validation.} The merit of any ZSL approach is usually evaluated empirically, since the assumptions cannot be validated absent samples in the unseen label class or knowledge of the transfer between the seen and unseen tasks. Once training has terminated and we have embeddings $\phi_{\hat w}$ and $\psi$, given test data $x^i$, we can compare the imputed labels obtained via
\begin{equation}
    \hat y(x_u) = \arg\min_{y \in Y_u} d(\psi \circ \phi_{\hat w}(x_u), y)
\end{equation}with labels $y_u$ in the validation set. The construction of this loss function \eqref{eq-ZSL-loss} is illustrated in Fig.~\ref{fig:diagram}.
 
\noindent{\bf Baseline.} ZSL methods differ by whether they learn both the visual and semantic embedding, only one, or none; by the method and the criterion used for learning. Since the unseen labels are never seen during training, the transfer of information from seen to unseen labels hinges on the {\em regularity} of the learned maps. For this reason, much of the recent work in ZSL aims to explore different regularization methods for the learned maps. The simplest case, which requires no regularization, is to assume that all the maps are linear: $\phi(x) = F x$ and $s(z) = V z$ for suitable matrices $F, V$ \cite{EZSL}. The results are not state-of-the-art (see Sect. \ref{sect-experiments}), but we nevertheless adopt this baseline and focus on regularizing, rather than the map $\psi$ directly, the spaces $Z$ and $S$, which is our key contribution. 

\section{Related Work }
\label{sec:related}

There are many variants for zero-shot learning (ZSL). The general formalism developed, along with the diagram in Fig. \ref{fig:diagram}, helps understanding the various approaches in relation to one another.

The problem of zero-shot learning dates back to the early days of visual recognition when the desire to transfer knowledge from painfully learned model to more general classes emerged \cite{OneShotLearningFei,Ullmanzero}. Early modern methods consistent with our approach include \cite{Lampert09,Norouzi} which can be described by a choice of fixed visual embedding $\phi$, semantic embedding $s$, and an imposed structure of the visual-semantic map (eq. (2) of \cite{Norouzi} in our notation)  
$$    \psi(z) = \sum_{i=1}^T z_i s(\hat y(z_i))$$
 where the sum is truncated at the $T$ largest elements of $z$, so nothing is learned. A particularly simple approach, which we adopt as baseline, is \cite{EZSL}, who assume that all the maps of interest are linear. In particular, they postulate
 $$ \psi(z) = V z. $$
 Although the map is linear, the domain and range where it is defined are not linear spaces (although their embedding space is). We adopt this choice and focus on smoothing the domain and range of the map.
 
Roughly speaking, zero shot learning methods can be classified into two main categories: inductive and transductive. In the inductive settings \cite{zhang2016zero,akata13cvpr,Lampart14,ARWLS15} which has dominated zero shot learning, the unseen classes are introduced one by one and decision about each unseen instance is made instantly once it is introduced. In the transductive setting \cite{cvprZ7Mapping_Paths,UnsupervisedZSL,Gopalan:TPAMI14,hangpinyoCGS16,ZSLwavelets,Song2018TransductiveUE} typically all unseen instances are processed simultaneously by constructing a graph where one exploits the underlying manifold structure, for example using the graph-based label propagation approach \cite{Gopalan:TPAMI14}. 
The problem of learning the graph Laplacian or the graph weights directly from given input data  has been recently addressed in a number of works \cite{kalofolias16,DongTFV16,EgilmezPO17,PavezO16,Lake10discoveringstructure}.
In the most general case, both the graph connectivity and the graph weights are unknown, in which case a common way to enforce smoothness is to use a regularizer which controls some level of sparsity. Perhaps the most widely used criterion is that the energy of the graph Laplacian computed from the graph signal at the vertices be small. Our approach is inspired from the isoperimetric problem, which is a classic problem in geometry. In Euclidean spaces the study of isoperimetric inequalities provides exact measures for general domains, while in Riemannian manifolds they provide some qualitative understanding of the geometry. Isoperimetric inequalities on manifolds were extended to graphs \cite{Chung:1997,Chung99highereigenvalues} where the analysis shares some similarities and intuition from the continuous settings. 
 
\section{Description of our approach}
\label{sec:description} 
 
 We select a fixed visual embedding $\phi$ consisting of a ResNet101 architecture trained on ImageNet using all classes, to map images $x$ onto a 2048-dimensional embedding $z = \phi(x)$. We assume $y_s^i \in Y_{s}$ is a subset of $n_s = 40$ to $150$ classes depending on the dataset: In  AwA \cite{Lampert09} there are $50$ classes, of which we consider $40$ as seen and sequester $n_u = 10$ as unseen. In CUB \cite{Cubdataset} there are $200$ classes, of which we consider $150$ as seen and the rest unseen. We exploit a fixed semantic map $s$ from text attributes, namely labels, onto a vector space $S = {\mathbb R}^M$ with dimension $M = 100$ (AwA) to $300$ (CUB), using Word2Vec \cite{MikolovSCCD13}:  The map $\psi: Z \rightarrow S$ is assumed linear, $\psi(z) = V z$, where $V$ is an $M\times 2048$ matrix, learned as in \cite{EZSL} using the seen dataset ${\cal D}_s$. To facilitate comparison to some algorithms we also use a VGGverydeep-19 on CUB rather than ResNet101.

At test time, given data $x_u^i$ with unseen labels, we compute the visual representation $z_u^i = \phi(x_u^i) \in {\mathbb R}^K$ and then semantic embeddings $s_u^i = V z_u^i$ for all $i = 1, \dots, N_u$. We construct a graph  ${\cal G} = (Z, W)$  with vertices\footnote{We abuse the notation to indicate with $Z$ the visual embedding space, the range of the function $\phi(X)$, which we assume to be a differentiable manifold, and the vertices of a discrete graph sampled from $Z$.}  $z_u^i \in Z$ and edges $\{w_{ij}\} = W$ that measure the affinities between embeddings $w_{ij} = \langle z_u^i, z_u^j\rangle$. ${\cal G}$ is a discrete representation of the smooth manifold $\phi(X) \subset Z$. The function $\psi$, restricted to ${\cal G}$, yields $s_u^i$, with range $\psi(Z)\subset S$, which we also assume to be a smooth manifold. In practice, because of the finite sampling and the nuisance variability in the descriptors, both the domain and range of $\psi$ are far from smooth. 

\noindent{\bf Key idea.} Rather than smoothing the map $\psi: {\cal G}\rightarrow S$,  we assume it is linear in the embedding space, and instead smooth both its domain and range. We seek a non-parametric deformation represented by changes in the connectivity matrix $W$ of the underlying graph, that minimizes the isoperimetric loss (IPL). This is a form of regularization which we introduce in the field of ZSL. The IPL measures the flow through a closed neighborhood relative to the area of its boundary. For two-dimensional (2-D) surfaces in 3-D, it is minimized when the neighborhood is a sphere. The IPL extends this intuition to higher dimensions. \\
{\bf Application to ZSL.} The result of our regularization is a new graph ${\cal G}'$, informed by the domain, range and map of the function $\psi$. We perform spectral clustering on ${\cal G}'$ to obtain a set of $n_u = |Y_u|$ clusters $\{c_1, \dots, c_{n_u}\}$. Each of these clusters can then be associated with a label in the unseen set $Y_u$. We do not need to know the association explicitly to evaluate our method. However, one could align the clusters to the semantic representation of the unknown labels if so desired.  
\footnote{ For instance, by finding a transformation $U$ that solves
\begin{equation}
    \min_{U\ge 0} \sum_{y_j \in Y_u} d\left(U s(y_j), 
    \frac{1}{|c_j|} \sum_{z^i_u \in c_j} {s^i_u}\right).
\end{equation}
If so desired, one could also add to the regularization procedure a term to align the clusters to the semantic representations of the unseen labels: 
\begin{equation}
    \sum_{z_i \in c_j} d( V z_i, U s(y_j)).
\end{equation}
We, however, skip this as the alignment issue is beyond our focus in this paper. In practice, we use the clustering of the regularized semantic attributes and the mapping found by using the Kuhn-Munkres algorithm, similar to \cite{Trigeorgis}. This does not have any impact on our method.} 

\noindent{\bf Results.} In general, there is no ``right'' regularizer, so we validate our approach empirically on the two most common datasets for ZSL, namely AwA and CUB. Compared to the current best methods that do not use any manual annotation, Zero-IPL reduces errors by 3.06\% on AwA1 (increased precision from 73.7\% to 76.03\%), and by 6.91\% (increased precision from 36.9\% to 39.45\%) on CUB.
Next, we describe the specific contribution, which is the smoothing of the graph-representation of $\psi$, in detail.

\section{Regularization}
\label{sec:regularization}
In this section we describe in more detail our graph smoothing based on the isoperimetric loss (IPL).

Our baseline gives us a graph ${\cal G}$ with weights ${w}_{ij}$ that we want to modify. We can think of these weights as ``noisy,'' and seek a way to regularize them, by exploiting also the function $\psi$ defined on $\cal G$ that yields semantic embeddings. Our regularization criterion is to achieve some level of compactness of bounded subsets: For a collection of subsets of the vertices with fixed size (corresponding to the volume of a subset) we want to find the subsets with the smallest size boundary. Why this might be a good criterion rests on classical differential geometry of Riemannian manifolds, where in the most basic case, the most compact manifold that encloses a fixed area with minimum size boundary is a circle. However, tools and concepts from classical differential geometry do not translate easily to graphs. Thus, we seek a technique that uses a key invariant, the isoperimetric dimension. It is transferred to the discrete setting, and we introduce the IPL as a way  to control smoothness in the graph. Our criterion, quantified by the isoperimetric gap, generalizes this bias towards compactness to more general sets. 

\subsection{Isoperimetric loss}

Let $B_{r}(\xi)$ be the ball around $\xi \in Z$  of radius $r$, that is the set of nodes within a distance $d_{\cal G}$ less than $r$. Let 
\begin{equation}
\mu_{i}^{(\xi)} = \sum_{i\sim j, \,\, d_{\cal G}(j,\xi) < d_{\cal G}(i,\xi)} w_{ij}
\label{offspring1}
\end{equation}be the flow from $i$ towards $\xi$, that is, the sum of weights of edges connecting $j$ with points closer to $\xi$. The geodesic flows $\mu_{r}^{(\xi)}$  are
\begin{equation}
\mu_{r}^{(\xi)} = \sum_{\underset{d_{\cal G}(i,\xi) =r } {i  }} \mu_{i}^{(\xi) }.
\label{offspring2}
\end{equation} Note that $\mu_{r}^{(\xi)}$ equals $\mu(\partial B_{r}(\xi))$ - the sum of all the edges that connect vertices in $B_{r}(\xi)$ and its complement in $Z$, where $\mu$ is a measure on the edges in the boundary $\partial B_{r}(\xi)$. 
Next we define the isoprimetric inequality.
\begin{mydef}\cite{Chung99highereigenvalues}
We say that a graph $\cal G$ has an isoperimetric dimension $\delta$ with a constant $c_{\delta}$, if, for every bounded subset $B_{r}(\xi)$ of $Z$,  the number of edges between $B_{r}(\xi)$ and the complement of $B_{r}(\xi)$, $Z\setminus B_{r}(\xi)$ satisfies 
\begin{equation}
 \, \mu(\partial B_{r}(\xi)) \geq c_{\delta} (\mu(B_{r}(\xi) ))^{1-\frac{1}{\delta}}
 \label{isopermetric_dimension_graph}
\end{equation}
where 
$\partial B_{r}(\xi)$ denotes the boundary of $B_{r}(\xi)$. 
\end{mydef} 

In our notation, we have that $ \partial B_{r}(\xi)  = \mu^{(\xi)}_{r}$. 

Next, we define the {\em isoperimetric gap} using the isoperimetric inequality above, which is the quantity to be minimized in the isoperiemtric loss: 
\begin{mydef}  The  {\em isoperimetric gap} is defined as 
\begin{equation}
\beta(\xi; \delta, W) \doteq  c_\delta \left(\sum_{i, j \in B_r(\xi)}w_{i,j} \right)^{1-\frac{1}{\delta}} - \mu^{(\xi)}_{r} 
\label{iso_gap}
\end{equation}
\end{mydef}
To minimize the gap we propose solving the following optimization problem:
\begin{equation}
\begin{matrix}
\underset{W\geq 0} {\mbox{min} }\,\sum_{\xi \in Z } \underset {z_u^i ,z_u^j \in B_{r}(\xi)}{\sum{f_{s_{u}^{i}, s_{u}^{j}}w_{i,j}} } + \lambda \beta(\xi; \delta, W) \\ 
\mbox{\rm s.t.\ }  
0\leq w_{i,j}\leq1 \, \forall \ i,j
\label{suggested_optimization}
\end{matrix}
\end{equation}
where $f_{s_{u}^{i}, s_{u}^{j}}$ is a function of the embedding distance between $s_{u}^{i}$ and $s_{u}^{j}$ (Specifically, ${f}: S \rightarrow \mathbb{R}$, where $S$ is the semantic embedding, and $f$ is the Euclidean distance; other choices are also possible) and $\lambda$ is a positive scalar tuning parameter. Note that the gap $\beta$ depends on $\delta$, the isoperimetric dimension, which is unknown, and will have to be approximated.  \\

{\textbf{Approximating the IPL}} To approximate the IPL loss, we elaborate on a spectral reduction of the isoperimetric loss, which provides a fast alternative to solving (\ref{suggested_optimization}) directly in the vertex domain. Approximating the IPL loss using a direct method to solve (\ref{suggested_optimization}) would entail approximating the isoperimetric dimension of the graph, which is challenging in general and even more for graphs constructed from noisy high dimensional data. Therefore, we choose to focus on a spectral reduction method.\\
\textbf{Spectral Reduction} We introduce a spectral reduction method for the isoperimetric inequalities, which reduces the isoperimteric gap directly in the spectral domain by using the vertex localization of Spectral Graph Wavelets (SGW) \cite{Hammond}. 
Specifically, we use the Spectral Graph Wavelet Transform of the semantic embedding space $s_{u}$ for each of the semantic dimensions $e$ of $s_{u}$. 

Let $s_{u}^{i}(e)$  be a component of $s_{u}^{i}$ in a fixed dimension $e$ 
 \begin{equation}
 \label{SGW_trans_ex}
 \begin{split}
 { {s}_{u}^{i}(e) }= \sum_{j=1}^{N} { s_{u}^{j}(e) } \sum_{k=0}^{r}{a_{k}\sum_{l=1}^{N}{\lambda_{l}^{k}\phi_{l}(j)\phi_{l}(i) } } = \\ \sum_{j=1}^{N}  s_{u}^{i}(e)\sum_{k=0}^{r} {a_{k}}(\mathbf{L}^{k})_{i,j}
 \end{split}
 \end{equation}a filtered signal of a fixed dimension $e$ of $s_{u}^{i}$, where $\phi_{l}$ is the corresponding eigenvector of the unnormalized Laplacian $\mathbf{L}$ associated with eigenvalue $ \lambda_{l}$. The coefficients $a_{k}$ are constants of a polynomial function and for a specific choice correspond to spectral graph wavelet (SGW) coefficients. Note that the terms  $\sum_{k=0}^{r} {a_{k}}(\mathbf{L}^{k})_{i,j} $ can be interpreted as the localized spectral transform of the graph around the ball $B_{r}(z_u^i(i))$, which vanishes for all $z_u^j \notin B_{r}(z_u^i)$. With the SGW transform, we employ a redundant representation with $r$ polynomials $\kappa_{t(k)}(\lambda)$, $1\leq k \leq r$, each is approximating a kernel function  localized in a frequency bands with corresponding scaling $t(k)$. Let $\delta_{i}$ be the impulse unit vector for the vertex $i$. Fixing a scale $t(k)$, the SGW coefficients $\psi_{i}^{t(k)}$ can be realized by $\psi_{i}^{t(k)} =   \sum_{j=1}^{N} \sum_{k=1}^{r} {a_{k}}(\mathbf{L}^{k})_{i,j} \delta_{i}$ and $\psi_{i,j}^{t(k)} =  \sum_{k=1}^{r} {a_{k}}(\mathbf{L}^{k})_{i,j} \delta_{i}$, where $\psi_{i,j}^{t(k)}$ is a scalar indicating the amount of diffusion propagated from vertex $i$ to $j$ in $B_{r}(z_u^i)$. Note that positive values $\psi_{i,j}^{t(k)}$ indicate that $j$ is informative about $i$, where small negative or zero values indicate insignificant influence with respect to the scale $t(k)$.  
Next choose the smallest $r_{0}$,  $1<r_{0}<r$ where we have $\psi_{i,j}^{t(r_{0})} \leq 0$ for all $j$, with the corresponding polynomial $\kappa_{t(r_{0})}(\lambda) $. Then, for all SGW coefficients $ { {s}_{u}^{i}(e) }$ in bands $k \geq r_{0}$, we annihilate all terms $ { {s}_{u}^{i}(e) }$, which has the effect of shrinking the boundaries of each ball around each vertex $i$ and thus reducing the isoperimetric loss directly in the spectral domain. Take the inverse transform to obtain the denoised signal and construct the new graph from the regularized semantic embedding space $\hat{s}_{u}$. 

 
The algorithm is summarized in pseudo code in Algorithm \ref{alg:buildtreeIPL}.
\begin{algorithm}[t]
\caption{Learning the Graph Connectivity Structure}
\label{alg:buildtreeIPL}
\begin{algorithmic}
\STATE{\textbf{ Process 1: Initialization: Embedding visual - semantic domains} } \\
\STATE{\textbf{Input}:  ${  z_{u} }= \left \{ z_{u}^{i} \right \}_{i=1}^{N_{u}}$,  $  s_{u} = \left \{ s_{u}^{i} \right \}_{i=1}^{N_{u}}$, $k$ nearest neighbor parameter, $r$ radius of the ball around $z_{u}^{i} $ }\\
\STATE{ \textbf{Step 1}: Construct $k$ nearest neighbor graph ${\cal G} = (Z,{{W}})$ from $\left \{ z_{u}^{i} \right \}_{i=1}^{N_{u}}$ (using cosine similarity)}. \\
\STATE{\textbf{Step 2}} Assign semantic attributes $s_{u}^{i}$ to its corresponding node $i$. \\
\STATE{\textbf{Output:} ${\cal G}= (Z,{W})$ }\\
\STATE{\textbf{Process 2}: \textbf{IPL Regularization} \\
\STATE{\textbf{Input:} graph ${\cal G}= (Z, {{W}})$, $ s_{u} = \left \{ s_{u}^{i} \right \}_{i=1}^{N_{u}}$.}\\
\STATE{\textbf{Step 1}: Construct the unnormalized Laplacian $\mathbf{L} = \mathbf{D} - W $ using ${\cal G}= (Z,{W})$.} Take the SGW transform (Eq. \ref{SGW_trans_ex}) for each dimension $i$ of of $s_{u}$.  }\\
\STATE{\textbf{Step 2}  Apply regularization using the spectral reduction of the IPL loss. }\\
\STATE{\textbf{Step 3} Construct a new graph using the regularized $\hat{s}_{u}$.  }\\
\STATE{\textbf{Output:} A new Semantic embedding graph ${\cal G}' = (S,{W_{s}})$}
\end{algorithmic}
\end{algorithm}

 \subsubsection{Clustering and validation in ZSL}

 We employ a standard procedure for spectral clustering as follows: the input is the graph  ${\cal G}' = (S,{W_{s}})$ obtained from applying the IPL algorithm, and the number of clusters, $n_{u}$. Construct the Laplacian matrix $\mathbf{L}_{s} $ using $W_{s}$ and compute the first $n_{u}$ eigenvectors of  $\mathbf{L}_{s} $. Letting  $\Phi \in \mathbb R^{ N_{u} \times  n_{u} } $ correspond to the first $n_{u}$ eigenvectors of $\mathbf{L}_{s} $ stacked in a matrix form, we cluster the vectors $\Phi_{i} \in \mathbb R^{ n_{u}}, i=1,..N_{u}$ corresponding to the rows of $\Phi$ into $n_{u} $ classes, using the k-means algorithm. Once $n_u$ clusters are found, we can associate each of them to a different unseen label. While it is not required that the semantic embedding of the unseen labels $s(y_u)$ correspond to the clusters in the same space, mapped from the visual embeddings, this alignment can be performed {\em post-hoc}. For the purpose of comparison, however, it is sufficient to perform the assignment by searching over permutations of the unknown labels. Since we have at most $50$ unseen labels in our experiments, this is not a bottleneck. More in general, one may consider introducing the alignment as part of the regularization, but this is beyond our scope in this paper.

\begin{figure*}[htb]
  \centering
 \includegraphics[width=16cm]{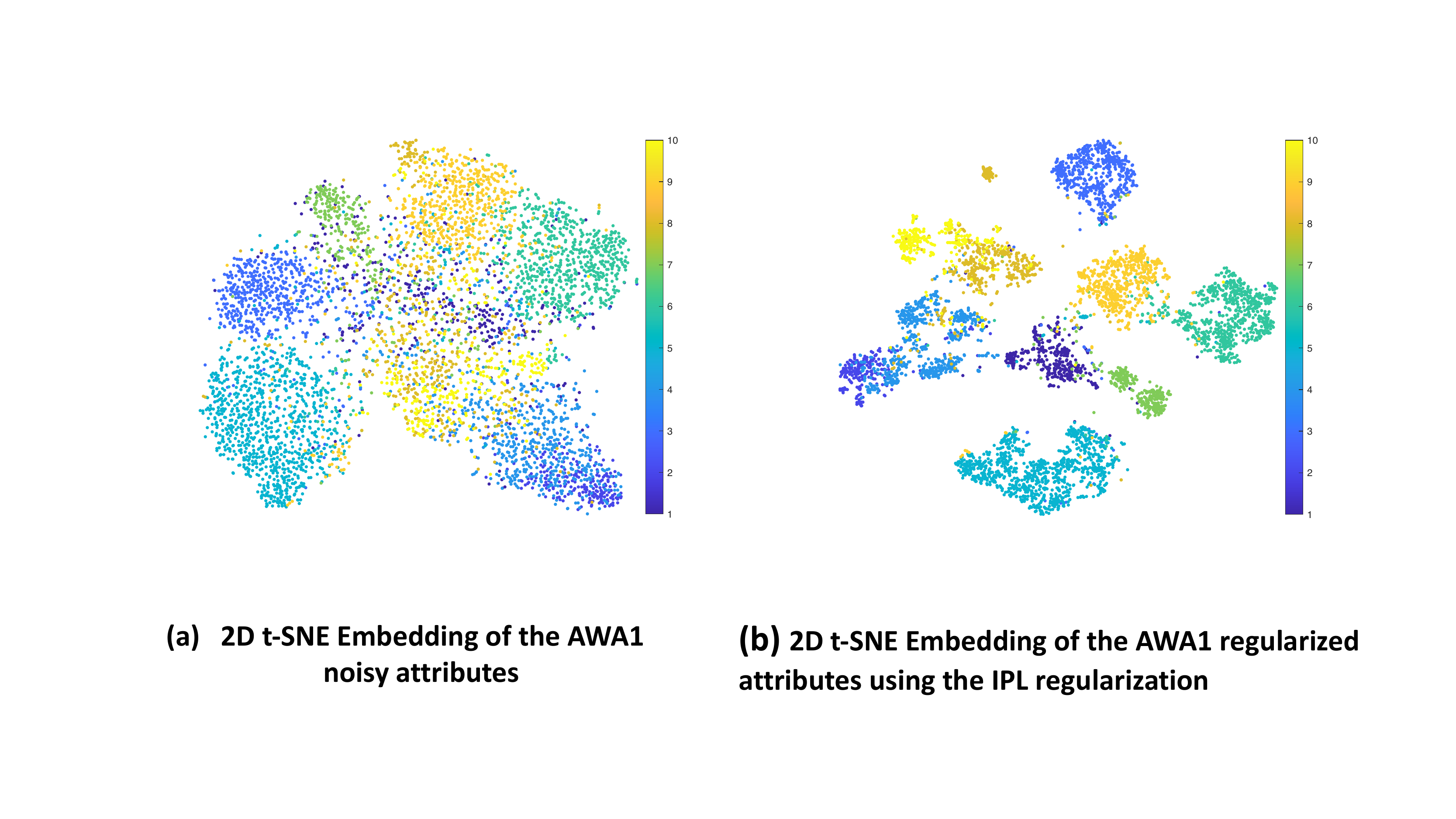}
\caption{An illustration using t-SNE embedding fot the AWA1 dataset comparing (a) noisy 2D embedding of the semantic attributes and (b) 2D embedding of the regularized semantic attributes using the proposed IPL regularization. Nodes with the same color correspond to the same class. The effect of the IPL regularization is clearly observed in (b), such that comparing to (a) the boundary of subsets, most in the same class, is typically shrinking and more compact}
\label{fig:t-sne embedding}
\end{figure*}


\section{Experimental Results}
\label{sect-experiments}

{\bf Experimental Settings.}
In the first set of experiments, we restrict our comparisons to approaches that are fully automated beyond the definition of the visual embedding (best performance is marked in boldface). In addition, we also report the evaluation of the state of the art methods that have access to embeddings of ground-truth semantic attributes. 

Using our approach we choose each component of our ZSL pipeline to be the simplest possible one, corresponding to the baseline \cite{EZSL}. A sanity check is whether our proposed regularization scheme improves over this baseline. Ideally, however, our method would take the baseline beyond the state-of-the-art.

 To test this hypothesis, we use the two most common benchmarks for ZSL, AwA and CUB. AwA (Animals with Attributes) consists of 30,745 images of 50 classes of animals and has a source/target split of 40 and 10 classes, respectively. In addition we test on the new released dataset AWA2 which consists of 37,322 images of 50 classes which is an extension of AwA (which will be refereed from now and on AwA1). AwA2 also has source/target of 40 and 10 classes respectively with a number of 7913 unseen testing classes. We used the proposed new splits for AwA1 and AwA2 \cite{goodbaduglycom}.

The CUB dataset contains 200 different bird classes, with 11,788 images in total.  We use the standard split \cite{hangpinyoCGS16} with 150 classes for training and 50 disjoint classes for testing  \cite{goodbaduglycom} which is employed in most automated based methods we compare to, while \cite{goodbaduglycom} also suggested a new split for the CUB dataset. Note that the CUB dataset is considered fine-grained, hence more challenging with both of the input features (visual and semantic) being very noisy. We present the evaluations in Tables \ref{tab:trans_compareAWA}, \ref{tab:CUB_dataset} and \ref{tab:compareGZSL} using methods which are either representative or competitive for ZSL using automated attributes including \cite{ZSLwavelets,UnsupervisedZSL,Latent_Embeddings,Devise,rahman2018unified,hangpinyoCGS16} as well as ones that used human annotation \cite{KernelZSL,Song2018TransductiveUE,zhang2017learning,relatinalcvpr18} for a more general overview.   \\\\
\textbf{Implementation details:} For all the splits of AwA and CUB datasets, we fix $k= 15$, $r=3$, and $k= 8$, $r=3$ for the $k$ nearest neighbor graph parameter and radius $r$ of the ball around each point, respectively. The edges ${w}_{ij}$ are chosen using the cosine similarity between the visual observations. \\

\begin{table}[t]
\begin{center}
\tabcolsep=0.001cm
\begin{tabular}{|l|c|c|c|c|}
\hline
Method/Data& AwA1  & AwA2  \\
\hline
{EZSL} \cite{EZSL} &58.2  &  58.6\\
\hline
 { SJE } \cite{ARWLS15}& 65.6  & 61.9 \\
\hline
 {ALE } \cite{Akata2016LabelEmbeddingFI}& 59.9  & 62.5\\
\hline
\textbf{LatEm} \cite{Latent_Embeddings} &50.8   & - \\
\hline
\textbf{DEVISE} \cite{Devise} & 50.4 &  -  & \\
\hline 
\textbf{SynC}  \cite{hangpinyoCGS16} & 58.6  & -  \\
\hline 
\textbf{CAPD} \cite{rahman2018unified} &64.73  & -  \\
\hline
{Kernel ZSL} \cite{KernelZSL}  & 71.0   & 70.51   \\
\hline 
{DEM \cite{zhang2017learning}  } &  68.4  & 67.1  \\
\hline
{RELATION NET}  \cite{relatinalcvpr18}  & 68.2  &  64.2  \\
\hline 
 {QFSL} \cite{Song2018TransductiveUE}  & -  &  79 \\
 \hline 
 {LisGAN} \cite{Li19Leveraging} & 70.6  &  -  \\
 \hline 
 GMN  \cite{Sariyildiz_2019_CVPR} & 82.5  &  -  \\
\hline
\textbf{MSMR}\cite{ZSLwavelets}	 & 73.7  &  72 \\
\hline
\textbf{Proposed} 	&  \textbf{76.03} 	 & \textbf{ 73.46}    \\
\hline
\end{tabular}
\end{center}
\caption{Mean average precision accuracy (top-1 in $ \%$) results using our method compared to the state of the art in ZSL on the AwA1 and AwA2 datasets. Best performance using automated semantic representation is marked in boldface. The evaluation for the state of the art methods which are using human semantic annotation is also presented.} 
\label{tab:trans_compareAWA}
\end{table}

\begin{table}[t]
\begin{center}
\tabcolsep=0.001cm
\begin{tabular}{|l|c|c|c|c|}
\hline
Method/Data& AwA1 & AwA2 & CUB\\
\hline
\cite{kalofolias16} &  58.8 & 55.2   & 33.45  \\
\hline
 \cite{GraphLearningA} & 66.6 & 66.7  & \textbf{39.6 } \\
\hline
\textbf{Proposed} 	& \textbf{76.03}    & 	\textbf{73.46 } & 39.4 \\
\hline
\end{tabular}
\end{center}
\caption{Mean average precision accuracy (top-1 in $ \%$) results using our method compared to the state of the art graph learning methods on the AwA1, AwA2 and CUB datasets. Best performance using automated semantic representation is marked in boldface.}
\label{tab:graphleanrning}
\end{table}

\begin{table}[t]
\begin{center}
\begin{tabular}{|l|c|c|}
\hline
Method/Data& CUB\\
\hline
\textbf{EZSL}  \cite{EZSL}&23.8 \\  
\hline
\textbf{SJE} \cite{ARWLS15}& 28.4 \\
\hline 
\textbf{LatEm} \cite{Latent_Embeddings} & 33.1\\
\hline
\textbf{Less Is more} \cite{CVPR16less_is_more}&29.2 \\
\hline
 {ALE} \cite{Akata2016LabelEmbeddingFI} & 54.9\\
 \hline
{Kernel ZSL} \cite{KernelZSL} &   57.1   \\
\hline 
{DEM} \cite{zhang2017learning}  &  51.7   \\
\hline
{RELATION NET}  \cite{relatinalcvpr18}  & 55.6  \\
\hline 
 {QFSL} \cite{Song2018TransductiveUE}  & 72.1  \\
 \hline 
 {LisGAN} \cite{Li19Leveraging}  &  58.8  \\
 \hline 
 {GMN}  \cite{Sariyildiz_2019_CVPR}    &  -  \\
 \hline 
\textbf{CAPD} \cite{rahman2018unified}  & 32.08 \\ 
\hline
\textbf{Multi-Cue ZSL} \cite{AkataMFS16} & 32.1 \\
\hline 
\textbf{DMaP}\cite{cvprZ7Mapping_Paths} & 30.34\\
\hline
\textbf{MSMR} \cite{ZSLwavelets}&  36.9   \\
\hline
\textbf{Proposed} &\textbf{ 39.45}   \\
\hline
\end{tabular}
\end{center}
\caption{Mean average precision accuracy (top-1 in $\%$) results using our method compared to the state of the art methods in zero shot learning on the CUB dataset using Word2Vec or other automated semantic representation. Methods using automated semantic representation are marked in boldface. The evaluation for the state of the art methods which are using human semantic annotation is also presented.} 
\label{tab:CUB_dataset}
\end{table}

\begin{table}
\begin{center}
\tabcolsep=0.001cm
\begin{tabular}{|l|c|c|c|c|}
\hline
Method/Data& AwA1  & AwA2  & CUB \\
\hline
{EZSL} \cite{EZSL} & 12.1   &  11.0 &  21.0 \\
\hline
 { SJE } \cite{ARWLS15}& 19.6  &  14.4 & 33.6 & \\
\hline
 {LatEm \cite{Latent_Embeddings} } & 13.3    &  20.0   &  24.0 \\
 \hline
 {ALE } \cite{Akata2016LabelEmbeddingFI} & 27.5 &  23.9 & 34.4 \\
\hline
{DEVISE } \cite{Devise}  &   22.4  &  27.8 &  32.8\\
\hline 
 {SynC } \cite{hangpinyoCGS16} & 16.2  & 18.0 & 19.8  \\
 \hline 
\textbf{DMaP} \cite{cvprZ7Mapping_Paths} & 6.44 & - & 2.07\\
\hline
\textbf{CAPD} \cite{rahman2018unified} &43.70   & -  &  31.6\\
\hline
{Kernel ZSL} \cite{KernelZSL}  & 29.8  & 30.8 & 35.1   \\
\hline 
{DEM \cite{zhang2017learning}  } &  47.3  &  45.1 & 13.6  \\
\hline
{RELATION NET}  \cite{relatinalcvpr18}  & 46.7  &  45.3 & 47.0  \\
\hline 
 {LisGAN} \cite{Li19Leveraging}  & 62.3  &  -  & 51.6  \\
 \hline 
   GMN  \cite{Sariyildiz_2019_CVPR}   & 74.8  &  -  & 65.0 \\
 \hline 
 {QFSL} \cite{Song2018TransductiveUE}  & -  &  77.4   & 73.2  \\
\hline
\textbf{Proposed}	&  \textbf{48.0} 	 & \textbf{ 49.2}   & \textbf{45.6 }\\
\hline
\end{tabular}
\end{center}
\caption{Comparison results in generalized ZSL on the AwA1, AwA2, and CUB data-sets. The harmonic mean is measured using the Mean average precision top-1$ \%$ accuracy using the unseen and seen classes. Methods using automated semantic representation are marked in boldface.} 
\label{tab:compareGZSL}
\end{table}

\textbf{Experimental results on the AwA1 and AwA2 datasets} using the new proposed splits \cite{goodbaduglycom} are shown in Table \ref{tab:trans_compareAWA}. Note that the new proposed AWA2 data-set is more challenging, as evident from the significant drop in performance compare to AwA for most of the state of the art methods. We also compare to state of the art methods which are employing human attributes (85 dimensional attribute vectors provided for each class  in \cite{Lampert09}). A ``-'' indicates that the performance of the method was not reported in the literature for the corresponding dataset. 
Mean average precision of the baseline is 58.6\%. We improve it to 
76.03\% and 73.46\% on the AwA1 and AwA2 datasets, respectively, by using our regularizer, taking the baseline past the state of the art, which is  
73.7 \% and 72\% using \cite{ZSLwavelets} on AwA1 and AwA2 respectively, reducing the error by 3.06 percentage points on AwA1. Note that among the most competitive state of the art methods which is also using automated attributes, \cite{ZSLwavelets} is using a much more complex and computationally heavy method.
Furthermore, for both AwA1 and AwA2, our method outperforms the state of the art methods which are using human attributes. Fig. \ref{fig:t-sne embedding} shows a comparison between the t-sne embedding of the noisy embedded semantic representation and the regularized semantic representation using IPL. 

\textbf{Experimental results on the CUB dataset} is the next benchmark we consider.  The baseline achieves a disappointing 23.8\% precision on CUB. Surprisingly, our regularizer takes it past the state-of-the-art automatic method (36.9\%), to 39.45\%, corresponding to an error decrease of over 6.9\%. The experimental results comparison on the CUB dataset is shown in Table \ref{tab:CUB_dataset}.  \\
\textbf{Influence of the $k$ nearest neighbor graph parameter}
We also provide additional experiments which test the influence of the $k$ nearest neighbor graph parameter. Changing the $k$ nearest neighbor graph parameter by 50\% for the AWA1, AWA2, and CUB datasets, results in a performance drop of less than 2.6, 8.77, and 2.02\%, respectively.


 \subsection{Comparison to Learning Graph Methods}
\label{Comp_Graph_Learning}

In addition to direct comparison to ZSL methods, since our approach uses a graph-based smoothing approach, one may wonder whether applying state-of-the-art graph learning methods one might also improve performance of ZSL. 

We focused on the state-or-the-art graph learning method \cite{GraphLearningA} and also \cite{kalofolias16} as representative of graph based learning. We use the same method and protocol to arrive at a graph, but replace our approach with \cite{GraphLearningA} \cite{kalofolias16} for evaluation in AwA1, AwA2, and CUB.  This experiment is comparable to the one reported in Table 1. While performance in CUB is comparable, our approach  outperforms \cite{GraphLearningA,kalofolias16} on the AwA benchmarks. 

\textbf{Experimental results on generalized ZSL (GZSL)}  We also compare our performance in the generalized zero shot learning setting. We follow the standard protocol of the generalized ZSL (GZSL)  \cite{Goodbadugcvpr} settings where the search space at evaluation time includes both the target and the source classes while the evaluation metric use the harmonic mean between while source and test data as the evaluation metrics. Thus, letting $Acc_{s}, Acc_{t}$ the mean class accuracy achieved for the source and target classes, respectively, the harmonic mean H is given by: 
\begin{equation}
H =  \frac{2 * Acc_{s} * Acc_{t} }{ Acc_{s}+ Acc_{t} }
\end{equation}

The settings of the GZSL is more challenging, as can be seen in the evaluation comparison (Table \ref{tab:compareGZSL}, for most methods the performance degrades significantly in comparison to the standard ZSL. We compare to the recent automated methods tested in the GZSL settings on the AwA1, AwA2 and CUB datasets (those which are available on GZSL and can scale to generalized ZSL settings). The experimental results summarized in Table \ref{tab:compareGZSL} show not only  improvement over method using automatic attributes  (error decrease of over 9.8\% and 44\% for the AWA1 and CUB datasets), but is also outperforming many of the recent state of the art methods which are using human annotation.  \\

\noindent{\bf Computational complexity:} The IPL with spectral reduction has a computational cost of $O(N_{u}K \log (N_{u}))$, which includes fast computation of the $k$ nearest neighbor graph using $k-d$ tree, the SGW transform which is $O(N_{u})$ for each dimension of the manifold for sparse graphs \cite{Hammond}, thus total complexity of $O(N_{u}K \log (N_{u}))$ for $N_{u}$ samples in $K$ dimensional space. Limitations includes processing very large graphs which consists more than millions of points. 
The execution time of our code implementation using Intel Core i7 7700 Quad-Core 3.6GHz with 64B Memory on the AWA1 dataset with 5685 points using $k=10$ nearest neighbor graph takes $\approx  21.9$ seconds for the initialization of the visual-semantic embedding space, and $\approx 44.8$ seconds for our IPL regularization. 


\section{Discussion}
\label{sec:discussion}

We have introduced the use of isoperimetric inequalities, known for centuries, into clustering in general, and zero-shot learning in particular. We use the isoperimetric loss to indirectly regularize a learned map from visual representations of data to their semantic embedding. Regularization is done by representing the domain of the map as a graph, the map as a graph signal, and regularizing the graph, obtaining another ``denoised'' graph where clustering is performed to reveal the unseen labels, once cluster-to-label association is performed. This regularization appears to be so effective as to take the simplest possible ZSL approach, where all maps are assumed linear, and improve it to beyond the current state-of-the-art for fully automatic ZSL approaches. 
Typical failure modes of our regularization and clustering algorithms are when the compactness values of the geodesic flows lie within a very wide range of different intervals corresponding to the bounded sets of the different input classes, 
which will result in incorrect estimation and detection of the geodesic flows and therefore ineffective regularization. 

Since our model is general, it could be used in conjunction with more sophisticated ZSL components, including those where the various maps are not linear, and learned jointly with regularization. 

\section{ Acknowledgments}

Research supported by ONR N00014-19-1-2229, N00014-19-1-2066, NSF grant DMS-1737770 and DARPA grant FA8750-18-2-0066.

\end{document}